%% file: main.tex
\documentclass{article}
\usepackage[final]{corl_2020} 

\usepackage[utf8]{inputenc}
\usepackage{graphicx}
\usepackage{amsmath}
\usepackage{color,colortbl}
\usepackage{tabularx}
\usepackage{wrapfig}

\definecolor{Blue}{rgb}{0,0,0.9}

\usepackage{algorithm}
\usepackage[noend]{algpseudocode}
\usepackage[normalem]{ulem}

\include{macros/text_macros}

\include{macros/referencing}

\include{macros/defines}

\title{Diverse Plausible Shape Completions from Ambiguous Depth Images}
\author{
Brad Saund and Dmitry Berenson\\
Robotics Institute\\
University of Michigan\\
\texttt{bsaund@umich.edu, dmitryb@umich.edu}}
\date{July 2020}

\begin{document}
\maketitle

\input{sections/abstract}
\input{sections/introduction}
\input{sections/related_work}

\input{sections/problem}
\input{sections/method}

\input{sections/evaluation}

\input{sections/experiments}

\input{sections/discussion}

\input{sections/conclusion}

\clearpage
\acknowledgments{This work was supported in part by NSF grant IIS-1750489 and by Toyota Research Institute (TRI). This article solely reflects the opinions of its authors and not TRI or any other Toyota entity. }


{\footnotesize
\bibliography{references}  
}

\clearpage
\appendix
\noindent{\huge \bfseries APPENDIX\par}
\input{appendix/ExtendedExperiments}

\end{document}

%% file: macros/text_macros.tex
\newcommand{\ourmethod}{\text{PSSNet}}

%% file: macros/referencing.tex
\newcommand{\figref}[1]{Fig.~\ref{#1}}
\newcommand{\algoref}[1]{Algorithm~\ref{#1}}
\newcommand{\algolineref}[1]{Line~\ref{#1}}



%% file: macros/defines.tex
\newcommand{\latent}{z}
\newcommand{\latenttraditional}{\latent^f}
\newcommand{\latentbox}{\latent^b}
\newcommand{\flowlatent}{\psi}

\newcommand{\mean}{\mu}
\newcommand{\logvar}{\text{logvar}}
\newcommand{\normal}{\mathcal{N}}
\newcommand{\normalpdf}{\varphi}
\newcommand{\networkinput}{x}
\newcommand{\networkoutput}{\tilde{y}}
\newcommand{\groundtruth}{y}
\newcommand{\loss}{L}
\newcommand{\reconstructionloss}{\loss^{\text{rec}}}
\newcommand{\flowloss}{\loss^{\text{flow}}}
\newcommand{\vaeloss}{\loss^{\text{VAE}}}

\newcommand{\shapecompletion}{f}
\newcommand{\plausibles}{\mathcal{P}}
\newcommand{\plausible}{\hat{\groundtruth}}
\newcommand{\sampledfrom}{\sim}
\newcommand{\numsamples}{n}
\newcommand{\completions}{\tilde{Y}}

\newcommand{\distance}{d}
\newcommand{\accuracy}{M_A}
\newcommand{\plausibility}{M_P}
\newcommand{\coverage}{M_C}
\newcommand{\quality}{M_{PD}}

\newcommand{\dataset}{D}
\newcommand{\depthimage}{Im}
\newcommand{\observationmodel}{obs}

\newcommand{\threshold}{\delta}

\definecolor{LightCyan}{rgb}{0.8,.9,1}
\newcommand{\chl}{\cellcolor{LightCyan}}

%% file: sections/abstract.tex

\begin{abstract}
We propose PSSNet, a network architecture for generating diverse plausible 3D reconstructions from a single 2.5D depth image. Existing methods tend to produce only small variations on a single shape, even when multiple shapes are consistent with an observation. To obtain diversity we alter a Variational Auto Encoder by providing a learned shape bounding box feature as side information during training. Since these features are known during training, we are able to add a supervised loss to the encoder and noiseless values to the decoder. To evaluate, we sample a set of completions from a network, construct a set of plausible shape matches for each test observation, and compare using our plausible diversity metric defined over sets of shapes. We perform experiments using Shapenet mugs and partially-occluded YCB objects and find that our method performs comparably in datasets with little ambiguity, and outperforms existing methods when many shapes plausibly fit an observed depth image. We demonstrate one use for PSSNet on a physical robot when grasping objects in occlusion and clutter.
\end{abstract}

%% file: sections/introduction.tex

\section{Introduction}

\begin{figure}
    \centering
    \includegraphics[width=0.99\textwidth]{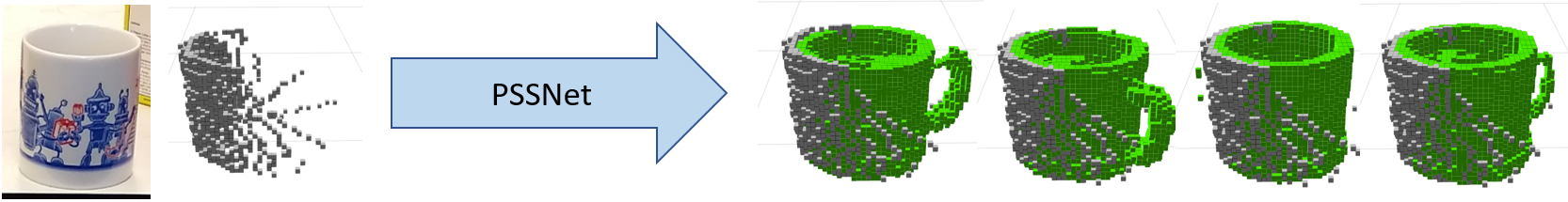}
    \caption{Our proposed $\ourmethod$ applied to a noisy segmented Kinect depth image of a mug produces multiple plausible reconstructions}
    \label{fig:live_completions}
\end{figure}

You look into a cabinet and see a coffee mug on the shelf.
Though you only observe the front of the shell you have a rich prior of shapes and so can infer the occluded structure of the mug.
Now suppose the handle is facing towards the back of the shelf, hidden from view.
You may imagine scenarios where the handle is on the left, on the right, straight back, or perhaps there no handle at all.
We propose a neural network architecture for generating these diverse samples over plausible completed shapes (Fig. \ref{fig:live_completions}).

More specifically, we generate a set of possible 3D shapes from a 2.5D depth image, such as that provided by a Kinect sensor.
There is inherent ambiguity in this process as it is impossible to know the true occupancy of occluded space. We thus seek an algorithm which produces a set of plausible 3D shape estimates from the observed data.


Broadly, researchers have attempted two approaches when inferring 3D structure from a 2.5D depth image.
Shape matching optimizes a model pose, potentially with uncertainty \citep{desingh2016physically, Peretroukhin2020}, thus requiring meshes of any potential object, limiting their ability to generalize.
Learning-based methods, such as Variational Auto Encoders (VAE) \citep{Wu2018LearningSP}, only require meshes during training and generate visually-pleasing shapes, but are optimized and evaluated on a single completion without consideration of other plausible completions.


Rather than operating on a single maximal-likelihood guess of the world, many robotics algorithms model and plan over a belief over worlds
\citep{saund2019}, thus we propose the Plausible Shape Sampling Network $\ourmethod$, capable of generating diverse shape completions when multiple plausible shapes could fit a depth image. Our \textit{key insight} is a restructuring of a Variational Auto Encoder to incorporate
shape-relevant
features during training.
We use a normalizing flow to map the pose and size of the shape's bounding box
into a portion of the latent space of the VAE.
During inference the network estimates a distribution over bounding boxes from which a specific box is sampled and used for reconstruction.

Evaluating the quality of our network presents a dilemma: how do we decide if shapes produced by our network are plausible when they differ from the ground truth?
We propose a non-learning method for generating plausible completions for a specific test dataset.

This paper makes the following contributions:
\begin{enumerate}
    \item A method to generate plausible completions for an evaluation dataset
    \item Metrics to evaluate plausible diversity of a black-box shape completer
    \item $\ourmethod$: A network for sampling diverse and plausible shape completions
\end{enumerate}

To validate our method, we perform experiments using mugs from shapenet \cite{shapenet2015} and all YCB objects \cite{YCB} which show that for ambiguous completions $\ourmethod$ produces diverse yet plausible samples, while baselines produce similar and poor quality completions.
Without ambiguity $\ourmethod$ still retains similar performance to baselines.
Finally, we construct physical robot scenarios of grasping objects in occlusion and clutter and show the diversity of $\ourmethod$ aids grasping.
Code and a video are available at \url{https://github.com/UM-ARM-Lab/probabilistic_shape_completion} and \url{https://youtu.be/mY6c8jeZVKU}

%% file: sections/related_work.tex

\section{Related Work}

\textbf{Shape Matching:} 
Robotics has studied the problem of inferring 3D structure from RGB and depth camera images for decades.
In the \textit{shape matching} variant the pose or configuration of a target shape is estimated from observations.
A classic but powerful non-learning approach uses the Iterative Closest Point (ICP) algorithm to align 
a target object with the observed pointcloud \citep{ICP, Yang20arxiv-teaser}, with some newer methods predicting a pose using neural networks \citep{Narayanan-2016-5537, PoseRBPF, hodan2020epos}.
Uncertainty can be modeled using discrete samples stored in a particle filter \citep{hausman2015active, Koval2015,Desingheaaw4523, desingh2016physically, desingh2019factored, liu2015table, saund2017datumpf}, where an observation model assigns a likelihood to each proposed shape based on the agreement with the observed depth image.
Researchers have hand-crafted likelihood models using sum of squared pixel depth distances \citep{desingh2016physically}, outlier rejection \citep{Narayanan-2016-5537}, gaussian per-pixel error \citep{Wuthrich2013}, and signed distance \citep{schmidt2014dart}.

Shape matching requires known meshes for objects, limiting the applicability in an unstructured novel world.
Our work uses shape matching to construct an evaluation dataset of plausible shapes and configurations for each given depth image.
Using ICP followed by an outlier rejection observation model, we generate plausible particles to evaluate how well $\ourmethod$ captures uncertainty.
$\ourmethod$ does not perform shape matching, nor require models outside of the training process.

\textbf{Shape Completion:} 
In \textit{shape completion} or \textit{shape reconstruction} the 3D structure is directly predicted from the camera observation.
Shape datasets such as shapenet \citep{shapenet2015} and YCB \citep{YCB} enable learning on sufficient examples to generate visually compelling results.
The most common network architecture learns an encoder to a feature space followed by a decoder to the shape output \citep{3DShapenets, choy20163d, Girdhar16b, Wu2018LearningSP, marrnet, michalkiewicz2020simple, Wen2019Pixel2MeshM3, xie2019pix2vox, Fan2017APS, Yu2020PointEG, Yang18, wu2016learning}.
In different variants the encoder may accept voxelgrids \citep{3DShapenets, marrnet, choy20163d, Yang18, Dai2017ShapeCU}, images \citep{Girdhar16b, xie2019pix2vox}, or point clouds \citep{yuan2018pcn, Fan2017APS}.
Similarly the decoder may produce voxelgrids \citep{3DShapenets, marrnet, choy20163d, Yang18, Dai2017ShapeCU, xie2019pix2vox}, point clouds \citep{yuan2018pcn, Fan2017APS}, meshes \citep{Wen2019Pixel2MeshM3}, octrees \citep{Riegler2017}, or implicit surfaces \cite{DeepSDF}.
Our proposed network encodes to and from voxelgrids, however we expect out contributions to be applicable to other approaches.

In these networks a reconstruction loss such as voxel-independent binary crossentropy guides the optimizer \citep{3DShapenets, choy20163d, Dai2017ShapeCU, wu2016learning}, which leads to averaging over possible shapes when there is ambiguity, producing ``blurry" completions.
Generative Adversarial Networks (GANs) \cite{goodfellow2014generative} penalize this averaging and are used to produce natural-looking 3D reconstructions \citep{wu2016learning, Wu2018LearningSP, Yang18}. 
We might hope that by employing VAEs with GANs we could sample substantially different yet plausible completions for a single input, yet this diversity has not been studied \citep{wu2016learning, Wu2018LearningSP, Yang18}.
In our experience VAE-GANs have resulted in visually pleasing samples with low diversity.

\textbf{Representing Bounding Box Uncertainty: } Our proposal for encouraging diversity involves explicitly training the feature space of a VAE to represent means and variances in properties such as position, orientation, and size.
The vector representation of these chosen features and their uncertainties must be representable and learnable by a neural network, which is a notorious challenge when representing rotations in SO(3).
While new rotation belief representations \citep{Peretroukhin2020} would be interesting to explore in our framework, we follow the approach of \citet{DOPE} and represent pose as a bounding box using 8 3-dimensional points.
However, the standard independent Gaussian prior of a VAE is a poor prior for boxes where we expect corner locations to be highly coupled.

Normalizing flows have become popular in image generation as a method to invertably and losslessly map the tightly coupled distribution of pixel values onto an independent Gaussian distribution \citep{dinh2014nice, realnvp, glow}.
However, normalizing flows have also been proposed to model posterior distributions of VAEs \citep{rezende2015variational, NVAE}.
We take a similar, but inverted, approach and learn a normalizing flow as a map from the distribution of bounding boxes to the same independent Gaussian distribution used in our VAE.

%% file: sections/problem.tex
\section{Problem Formulation and Metrics} \label{sec:problem}


We assume a dataset of pairs $(\networkinput, \groundtruth)$ where $\networkinput$ is the two voxelgrids (known occupied, known free) for voxelized shape $\groundtruth$.
In this work we refer to an \textit{object} as a mesh at an unspecified pose and a \textit{shape} as a voxelgrid produced by an object at a specific pose.
We assume that for each $\networkinput$ there is given a set of plausible completions $\plausibles(\networkinput)$.
We desire a non-deterministic function $\networkoutput_i \sampledfrom \shapecompletion(\networkinput)$ where $\networkoutput_i$ is a voxelgrid called a \textit{completion} of $\networkinput$. 
Drawing $\numsamples$ samples from $\shapecompletion(\networkinput)$ gives a set of completions $\completions_\networkinput = \{\networkoutput_1, ..., \networkoutput_\numsamples\}$.
Let $\distance(\groundtruth_1, \groundtruth_2)$ be a distance function between two voxelgrids (e.g. Chamfer Distance). 
We define the Best Accuracy as $\accuracy(\networkinput) = \min_{\networkoutput_i \in \completions_\networkinput} \distance(\networkoutput_i, \groundtruth) $.
For a given $(\networkinput, \groundtruth)$ pair in our test dataset we additionally evaluate the quality of $\shapecompletion$ using 3 criteria:

\textbf{1. The coverage of plausible completions:}
\begin{align}
    \coverage(\networkinput) &= \frac{1}{|\plausibles(\networkinput)|} \sum_{\plausible \in \plausibles(\networkinput)} \min_{\networkoutput_i \in \completions_\networkinput} \distance(\networkoutput_i, \plausible)
    \label{eq:coverage}
\end{align}

\textbf{2. The average plausibility of completions generated by $\shapecompletion$:}
\begin{align}
    \plausibility(\networkinput) &= \frac{1}{|\completions_\networkinput|} \sum_{\networkoutput_i \in \completions_\networkinput} \min_{\plausible \in \plausibles(\networkinput)} \distance(\networkoutput_i, \plausible)
    \label{eq:plausibility}
\end{align}

\textbf{3. The Plausible Diversity:}
\begin{align}
    \quality &= \coverage + \plausibility \label{eq:quality}
\end{align}

$\accuracy$ is most similar to metrics used in previous work and is also not dependent on construction of $\plausibles$.
$\coverage$ penalizes plausible shapes that are not generated by $\shapecompletion$, whereas $\plausibility$ penalizes network samples that are far from $\plausibles$.
We want to generate diverse samples that are plausible, thus we seek an $\shapecompletion$ that achieves lowest $\quality$, which is the chamfer distance between the sets $\plausibles$ and $\completions$.





%% file: sections/method.tex

\section{Method} \label{sec:method}
\subsection{Plausible Shape Sampling}

\begin{figure}
    \centering
    \includegraphics[width=0.99\textwidth]{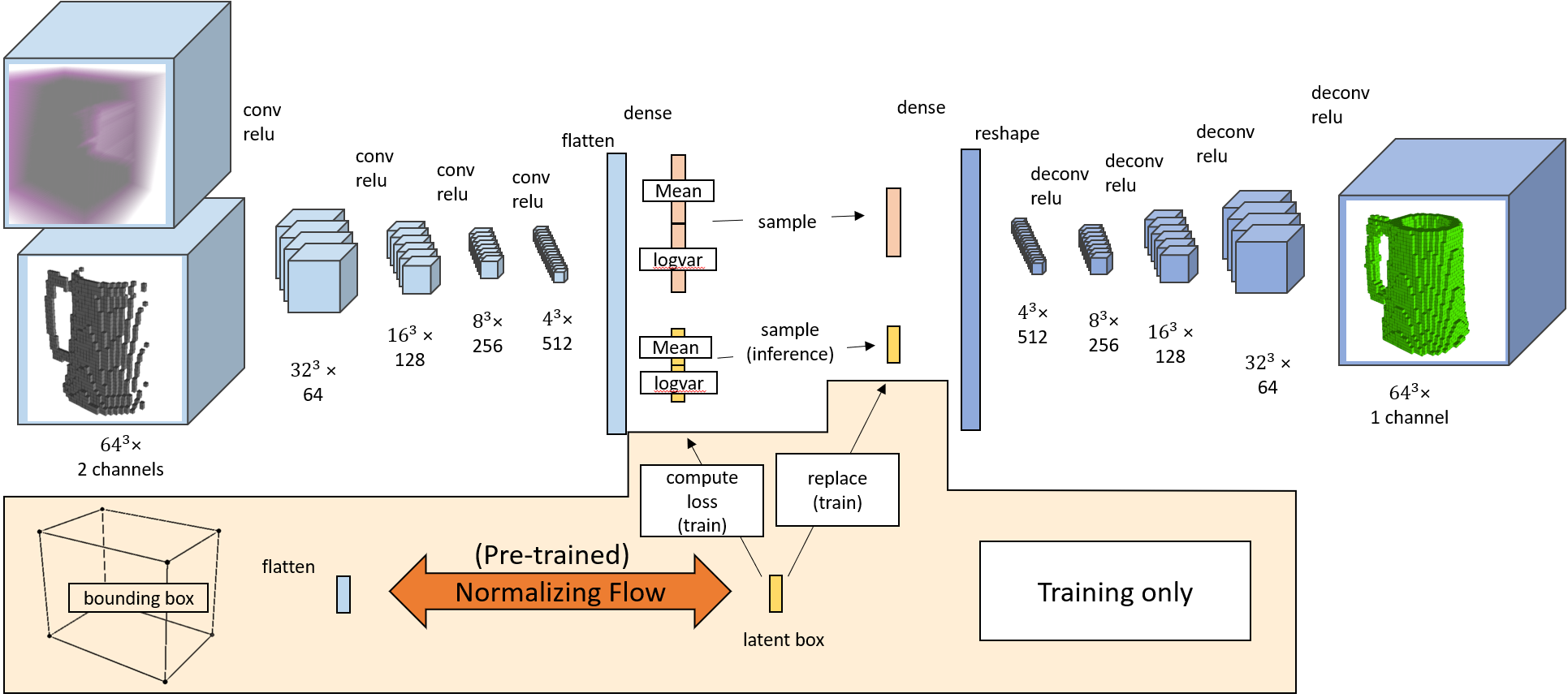}
    \caption{Our network, $\ourmethod$, has the structure of a VAE during inference. During training we separate the latent space into typical learned VAE features and ``latent box" feature produced by a learned normalizing flow applied to the ground truth bounding box. These latent box features are used both as a loss on the encoder prediction and as input to the decoder during training.}
    \label{fig:network}
\end{figure}

Our Plausible Shape Sampling Network, $\ourmethod$, is an adaptation of a variational auto encoder (VAE).
During inference $\ourmethod$ exactly follows a VAE, with an encoder that predicts a latent mean and variance from which a latent vector is sampled, and a decoder that produces a 3D voxelgrid from this latent vector.
During training $\ourmethod$ differs from a VAE by replacing a portion of the latent space with a learned representation of an additional input.

Our training data starts with a set of mesh objects at a single pose.
For each object we compute an axis-aligned bounding box. 
We then augment the dataset by applying rotations and translations to each object and bounding box.
Finally, we compute the voxelized shape $\groundtruth$ and the known-free and known-occupied voxels $\networkinput$ from a fixed view.

We train a normalizing flow on the bounding boxes of the training dataset with a Gaussian prior $\normal(0, 1)$.
Each bounding box consists of 8 points, thus this flow maps from a 24 dimensional ``box" space to a 24 dimensional ``latent box" space $\flowlatent$. 
The flow consists of 8 RealNVP networks \citep{realnvp}, each with 2 hidden layers of size 512. During training batch normalization is performed between every other RealNVP network.

We then use this flow in training $\ourmethod$ (\figref{fig:network}). 
The encoder takes as input $\networkinput$ the known-occupied and the known-free $64^3$ voxelgrids. 
$2 \times 2 \times 2$ convolutions with a stride of 2 and relu activation are applied 4 times sequentially using [64, 128, 256, 512] channels. 
The output is densely connected to a 200D latent-mean and 200D latent log-variance.
During inference the network is identical to a VAE, and thus a latent vector is sampled from this mean and log-variance.
The decoder inverts the structure of the encoder, with a dense layer reshaped into a 4x4x4x512 tensor followed by ``deconvolution", or convolution-transpose layers again with a stride of 2. 
The output of the decoder, $\networkoutput$, is a $64^3$ voxelgrid that represents the probability of occupancy for each voxel, independently. 
We threshold this voxelgrid at 0.5 to produce a binary occupancy.

During training, $\ourmethod$ differs from a VAE during the latent space sampling.
The latent space is partitioned into two vectors: $\latenttraditional$ and the 24 dimensional latent box space $\latentbox$.
During training $\latentbox$ is replaced by $\flowlatent$, the latent box produced by the normalizing flow applied to the bounding box, thus $\latentbox$ has no effect on the final voxelgrid produced.
A loss term $\flowloss$ rewards the log-likelihood of $\flowlatent$ given the latent mean $\latentbox_\mean$ and variance $\latentbox_\logvar$ produced by the encoder.
Additional loss terms for binary cross-entropy reconstruction loss $\reconstructionloss$ and $\vaeloss$ form the Monte Carlo estimate of the Evidence Lower Bound (ELBO) \citep{kingma2014autoencoding} as applied to shape completion \citep{Yang18, Wu2018LearningSP}.
With $N$ as the total number of voxels ($64^3$), $\groundtruth[i]$ as the target value $\{0, 1\}$ of the $i$th voxel, and $\normalpdf(\mu, \sigma_{logvar})$ is the probability density at $\mu$ of a Gaussian with log-variance $\sigma_{logvar}$.

\begin{align}
    \reconstructionloss &= p(y | \latent) 
    &&=\frac{1}{N}\sum_{i=1}^N - \groundtruth[i] \log(\networkoutput[i]) - (1 - \groundtruth[i]) \log(\networkoutput[i])\\
    \vaeloss &= \log(p(\latenttraditional)) - \log(p(\latenttraditional|x)) 
    &&= \log(\normalpdf(\latenttraditional, 0)) - \log(\normalpdf(\latenttraditional - \latenttraditional_\mean, \latenttraditional_\logvar))\\
    \flowloss &= \log(p(\flowlatent | \latentbox_\mean, \latentbox_\logvar)) 
    &&= \log(\normalpdf(\flowlatent - \latentbox_\mean, \latentbox_\logvar))
\end{align}

%% file: sections/evaluation.tex

\subsection{Quantifying Plausibility} \label{sec:evaluation}

Many shape completion methods evaluate results using the metric $\distance(\shapecompletion(\networkinput), \groundtruth)$, which may be appropriate if the ground truth shape is unambiguous given the view from the depth camera.
However, given two different shapes $\groundtruth_1, \groundtruth_2$ in the dataset with similar corresponding depth camera image $\networkinput_1 \approx \networkinput_2$ it is unreasonable to expect $\shapecompletion$ to always generate the correct output. 
Furthermore, for our application we desire $\shapecompletion$ to output diverse yet plausible shapes.

We propose two criteria to define some $\groundtruth_j$ as a plausible completion of $\networkinput_i$:
\begin{itemize}
    \item Observing $\networkinput_i$ given $\groundtruth_j$ must be sufficiently likely given a camera observation model
    \item The object represented by $\groundtruth_j$ is in the test database, possibly with a different pose
\end{itemize}

To address the first criterion we define an observation model $\observationmodel(\networkinput, \groundtruth)$ as the likelihood of observing the depth image of the 2.5D view $\depthimage(\networkinput)$, given that the true occupied voxels are $\groundtruth$.
Similar work uses the sum-of-squared depth differences of $\depthimage(\networkinput) - \depthimage(\groundtruth)$ \citep{desingh2016physically}, yet we find this model is not sufficiently discriminative.
On the other hand, applying a Gaussian belief to each pixel independently \citep{Wuthrich2013} is far too discrimative, as a single pixel can alter the likelihood by orders of magnitude.
We have had the most success with an outlier rejection model \citep{Narayanan-2016-5537}.

We define our $\observationmodel(\networkinput, \groundtruth)$ as a binary likelihood in \algoref{alg:observationmodel}, indicating if $\networkinput$ is or is not plausible.
We first compute a mask of unreliable depth pixels as any pixel in $\depthimage(\groundtruth)$ with gradient greater than some threshold $\threshold$, and inflate this mask by one pixel (\algolineref{alg:observationmodel:mask}).
We accept $\networkinput$ as a plausible depth image of $\groundtruth$ if every reliable pixel of $||\depthimage(\networkinput) - \depthimage(\groundtruth)||$ is below $\threshold = 4$cm.
Depending on sensor noise it may be appropriate to allow some outliers.
We deem certain pixels in the depth image $\depthimage(\groundtruth)$ ``unreliable" if they are at the boundary of shape, as discretization approximations due to pixelization may assign a vastly different depth value due to a slight translation orthogonal to the camera. 
We see this effect on physical hardware such as a Kinect as depth values near the boundary of shapes are sometimes far too large, causing points to trail off into the background.







With $\observationmodel$ now defined, we generate candidate shapes using objects from the test dataset $\dataset_{TEST}$.
Uniformly sampling poses and objects is infeasibly inefficient, as the vast majority of samples are not plausible.
As in some particle filter approaches \citep{Klingensmith-implicit-MPF}, we sample candidate states and project these onto a manifold of states more likely to be plausible. 
\algoref{alg:plausibles} describes our approach. 
For each $(\networkinput_i, \groundtruth_i) \in \dataset_{TEST}$ we attempt to create a plausible completion using every element $(\networkinput_j, \groundtruth_j) \in \dataset_{TEST}$.
We find a transformation $T$ to align the 2.5D voxelgrids $\networkinput_j$ to $\networkinput_i$ using $ICP$ \citep{PCL} (\algolineref{alg:plausibles:ICP}).
We then check if the observation is plausible given this aligned shape. 

\begin{minipage}{0.57\textwidth}
\begin{algorithm}[H]
  \caption{Observation Plausible: $\observationmodel(\networkinput, \groundtruth)$}
  \label{alg:observationmodel}
  \begin{algorithmic}[1]
    \State obs\_image = $\depthimage(\networkinput)$
    \State exp\_image = $\depthimage(\groundtruth)$
    \State mask = ComputeUnreliable(expected\_image) \label{alg:observationmodel:mask}
    \For{each pixel index $i$ not in mask}
      \If{$||$obs\_image$[i]$ - exp\_image$[i]||$  $> \threshold$}
        \State \textbf{return} False
      \EndIf
    \EndFor
    \State \textbf{return} True
  \end{algorithmic}
\end{algorithm}
\end{minipage}
\begin{minipage}{0.4\textwidth}
\begin{algorithm}[H]
  \caption{Compute Plausibles($\networkinput_i$)} \label{alg:plausibles}
  \begin{algorithmic}[1]
    \State $\plausibles(\networkinput_i) = \emptyset$
    \For{$(\networkinput_j, \groundtruth_j) \in \dataset_{TEST}$}
    \State T = $ICP(\networkinput_j, \networkinput_i)$ \label{alg:plausibles:ICP}
    
    \If{$\observationmodel(\networkinput_i | T \groundtruth_j)$} 
    \State $\plausibles(\networkinput_i) = \plausibles(\networkinput_i) \cup T \groundtruth_j$
    \EndIf
    \EndFor
    \State \textbf{return} $\plausibles(\networkinput_i)$
  \end{algorithmic}
\end{algorithm}
\end{minipage}

%% file: sections/experiments.tex

\section{Experiments}
\label{sec:experiments}

We present quantitative and qualitative results demonstrating that for non-ambiguous completions $\ourmethod$ performs on par with existing methods, and that when there is ambiguity $\ourmethod$ performs better.
We created datasets from shapenet \citep{shapenet2015} and YCB \citep{YCB} such that 2.5D views could have multiple consistent completions.
We trained $\ourmethod$ as described above as well as a VAE, a VAE with GAN loss similar to \citep{wu2016learning}, and 3D-rec-GAN++ (without super-resolution layers) \cite{Yang18}, with networks accepting and producing voxelgrids of size $64^3$.
We constructed plausible completions $\plausibles$ for each $\networkinput$ in our test dataset and evaluated our metrics (Section \ref{sec:problem}) using $\distance(y_1, y_2)$, as chamfer distance between voxelgrids converted to pointclouds, as it is a common metric of shape completion quality \citep{Wu2018LearningSP}.


\begin{figure}
    \centering
    \includegraphics[width=0.99\textwidth]{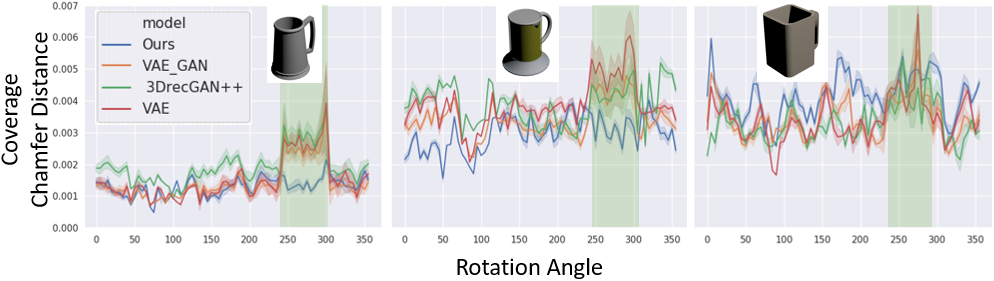}
    \caption{Coverage (Eq. \ref{eq:coverage}) of various methods for shapenet mugs at different rotation angles. Rotations where the mug handle is occluded are highlighted.}
    \label{fig:shapenet_coverage}
\end{figure}
\begin{figure}
    \centering
    \includegraphics[width=0.99\textwidth]{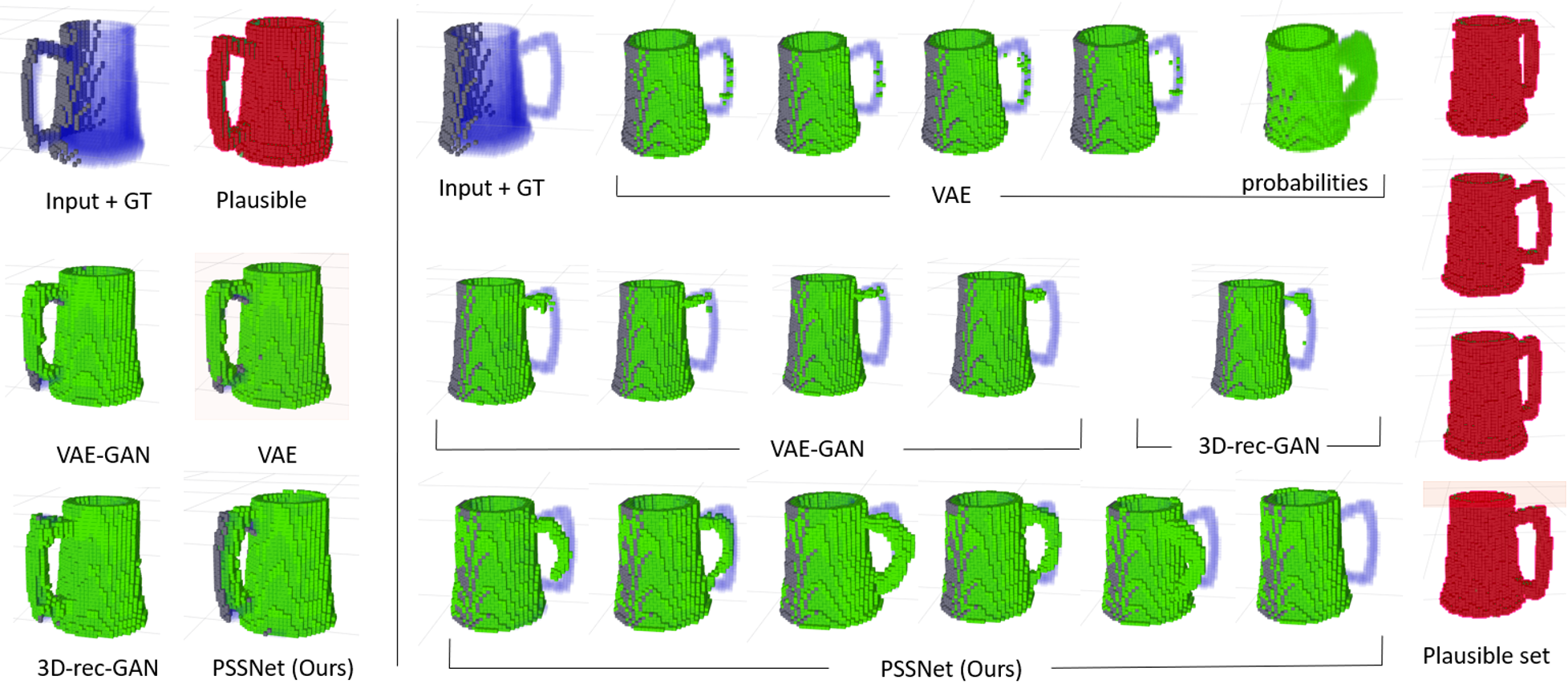}
    \caption{Completions (green) of a mug are sampled from the visible 2.5D view (grey). When the handle is visible (left) all methods produce similar mugs close to the ground truth (GT) (blue). When the handle is occluded (right) sampling from $\ourmethod$ yields mugs with different styles of handles in different orientations, with similar variation seen in the plausible set (4 shapes shown).}
    \label{fig:shapenet_completions}
\end{figure}

\begin{table}
    \centering
    \begin{tabular}{|c|c|c|c|c||c|c|c|c|}
    
    \hline
         & \multicolumn{4}{|c||}{Shapenet: all mugs}  & \multicolumn{4}{c|}{Shapenet: occluded handle}\\
        \hline
         &  \shortstack{best \\ acc}
         & \shortstack{coverage \\ of $\plausibles$}  
         & \shortstack{avg. \\ plaus} 
         & \shortstack{plausible \\ diversity}
         &  \shortstack{best \\ acc} 
         & \shortstack{coverage \\ of $\plausibles$} 
         & \shortstack{avg. \\ plaus}  
         & \shortstack{plausible \\ diversity} \\

        \hline
        $\ourmethod$ (ours)&\chl 1.9 &\chl2.9 &    2.0 &\chl4.9    &  \chl2.0 &\chl3.0 &\chl1.9 & \chl5.0    \\
        VAE        &     2.0 &    3.3 &    1.9 &    5.2    &      2.7 &    3.6 &    2.2 &    5.8    \\
        3D-rec-GAN &     2.2 &    3.6 &    2.0 &    5.6    &      3.0 &    3.9 &    2.3 &    6.2    \\
        VAE-GAN    &     2.0 &    3.3 &    1.9 &    5.2    &      2.8 &    3.7 &    2.3 &    5.9    \\
        \hline
    
    \hline
    \hline
         & \multicolumn{4}{|c||}{
         YCB: 30 pixel wide slit}  & \multicolumn{4}{c|}{YCB: 6 pixel narrow slit}\\
        \hline
$\ourmethod$ (ours)       &     1.3 &\chl1.7 &    3.2 &\chl4.8    &  \chl2.3 &\chl4.5 &    4.4 &\chl8.9    \\
VAE        &     1.5 &    3.1 &    1.8 &    4.9    &      3.0 &    7.8 &    2.8 &   10.6    \\
3D-rec-GAN & \chl1.2 &    3.6 &\chl1.2 &\chl4.8    &      4.6 &    9.6 &    2.9 &   12.4    \\
VAE-GAN    &     1.3 &    3.3 &    1.6 &    5.0    &      3.1 &    7.9 &\chl2.7 &   10.6    \\
        \hline
    \end{tabular}
    \caption{Best sample accuracy, Coverage of the plausible set, Average sample plausibility and Plausible diversity in mm. $\ourmethod$ performs best relatively in ``Shapenet: occluded handle" and ``YCB: narrow slit", as in these datasets there is ambiguity in the full shape given the partial view.
    }
    \label{tab:data}
\end{table}

\textbf{Shapenet Mugs:} 
Using the \textit{mugs} category from shapenet we constructed a dataset of 177 train and 37 test meshes. 
We rotated each mesh and associated bounded box in 5 degree increments about the vertical axis and voxelized using binvox \cite{binvoxcode}, creating 12744 train and 2664 test shapes.
For approximately 1/5 of rotations, the handle is completely occluded from the 2.5D view.

We display the coverage metric for three of these shapes in \figref{fig:shapenet_coverage}.
The left and middle mugs have a typical handle and when the handle is visible all methods obtain similar coverage.
When the handle is occluded other methods perform far worse on $\coverage$, meaning there are plausible completions that significantly differ from any samples produced by the network.
$\ourmethod$ retains similar coverage even in these occluded regions.
The right mug is square and unlike mugs in the training dataset, and the chamfer distance reconstruction error is dominated by the mug body reconstruction.

We visualize samples in \figref{fig:shapenet_completions} and qualitatively observe the same trends.
When visible, all methods accurately reconstruct the mug handle, but when occluded other methods tend to average over plausible mugs and produce poor and non-diverse samples.
For the 7 mugs from $\ourmethod$ the handles vary in orientation and style while remaining in the occluded region.
We find $\ourmethod$ generates these diverse plausible handles for many but not all mugs.
Qualitatively, we observe similar behavior for $\ourmethod$ with live Kinect depth images using a hard-coded segmentation of a mug (\figref{fig:live_completions}).

\begin{figure}
    \centering
    \includegraphics[width=0.99\textwidth]{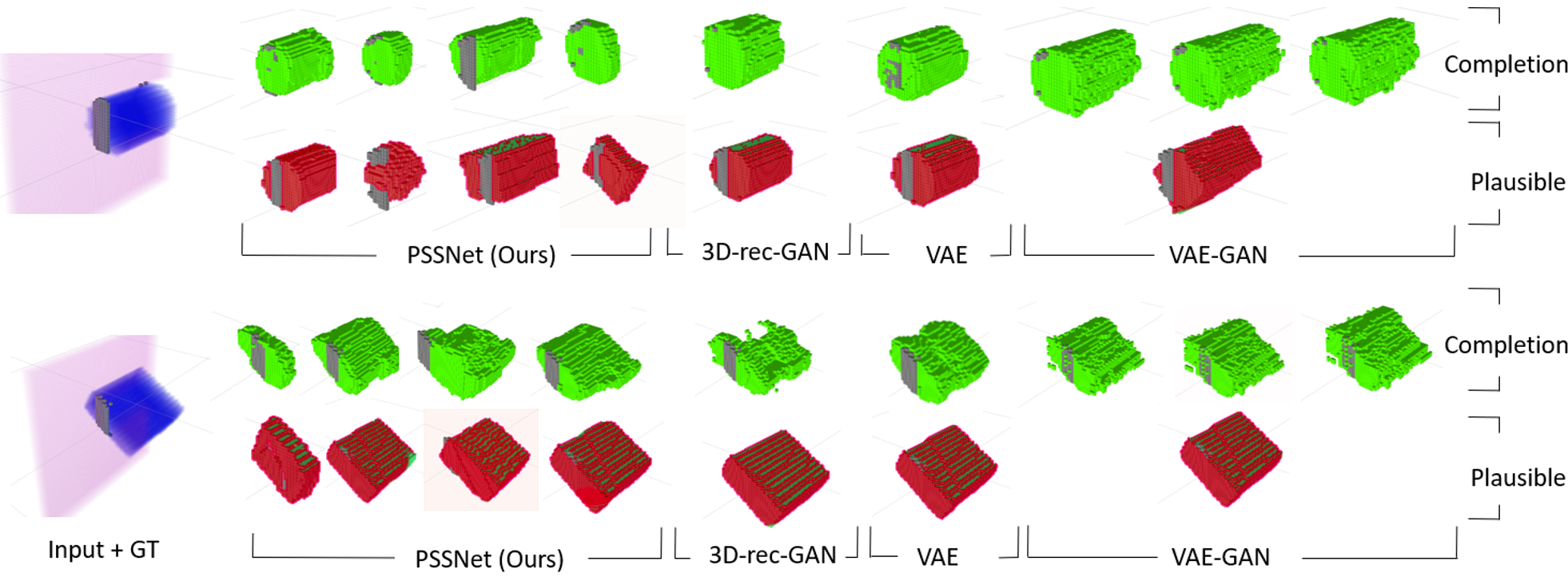}
    \caption{Completions of YCB objects as viewed through a 6 pixel narrow slit with the nearest plausible shape shown for each network sample. $\ourmethod$ generates diverse samples where other networks generate only small variations on the same sample.}
    \label{fig:ycb_completions}
\end{figure}

\textbf{YCB with slit occlusion:}
We constructed a training dataset by applying a total of 24 rotations about the vertical and a horizontal axis for each YCB object.
During training we occlude left and right portions of the depth image to simulate viewing the object through a vertical slit.
We randomly translate the YCB shape and then randomly select a slit of width 5 to 30 pixels (1 pixel $\approx$ 0.6cm) and randomly place this slit so that the target object is visible in at least 5 columns of the image. 
A full 2.5D view of any YCB object leaves little ambiguity, and this slit simulates viewing occluded objects in a cluttered scene.


We construct two test datasets for a subset of the YCB objects by using the same set of rotations but fix the translations and fix slit widths to 6 and 30 pixels.
For each fixed slit width we construct a separate $\plausibles$ by fitting (Alg. \ref{alg:plausibles}) each test shape at each orientation and each translation along the slit in 2 pixel increments.
6 pixels is a small portion of each object, thus in this dataset different objects with many different translations tend to match each $\networkinput$.
The 30 pixel slit captures most of the object, so there is little ambiguity as to the 3D shape.
We visualize completions in \figref{fig:ycb_completions}.

Metrics averaged over all test datasets are shown in Table \ref{tab:data}.
$\ourmethod$ consistently provides the best coverage and comparably in plausible diversity for the datasets with lower ambiguity and outperforms baselines for datasets with greater ambiguity.

\begin{figure}
    \centering
    \includegraphics[width=0.99\linewidth]{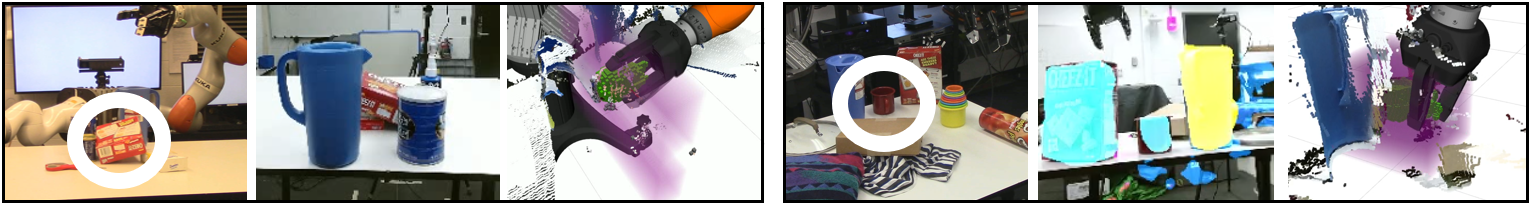}
    \caption{Robot scenarios for grasping Cheez-it box (left) and the mug (right).
    From left to right: The scene, the robot's view of the scene, and a grasp attempt.}
    \label{fig:robot}
\end{figure}

\textbf{Physical Robot:}
We constructed two scenarios on a physical robot, shown in \figref{fig:robot} and the accompanying video, where a grasp is chosen based on completions of a mug and the YCB Cheez-it box.
Consistent with the simulation experiments we found the baseline methods did not produce reasonable handles for the mug and thus grasps failed, while $\ourmethod$ produced multiple plausible handles leading to a successful grasp. 
Similarly, for the Cheez-it box viewed throw a narrow slit formed from other clutter, baseline methods produced nearly no variation about an incorrect completion, thus the attempted top grasp was not successful.
$\ourmethod$ produced a variety of completions, some similar to the Cheez-it box and some more similar to other YCB objects. 
With this ambiguity, the robot executed a side grasp that would capture many of the different possibilities, and successfully grasped the Cheez-it box.
Details are further described in the appendix \ref{sec:physical}.

%% file: sections/discussion.tex
\section{Discussion}

We achieve our goal of creating a network that generates diverse samples, while other networks generate only small variations on a single completion.
$\ourmethod$, however, performs worse on $\plausibility$, indicating that either $\ourmethod$ sometimes produces poor quality samples, or that $\plausibles$ lacks some plausible completions.
Subjectively, we see both cases.
Given a larger set of test shapes, $\plausibles$ would contain more shapes, and likely $\plausibility$ would improve.
Below we discuss what we see as the main advantages and limitations of our design choices for $\ourmethod$ and the plausible set:

\textbf{Feature replacement:}
The partial feature replacement in the latent space of the VAE allows proper credit assignment between the encoder and decoder during training of ambiguous samples.
For inputs where the reconstruction is inherently ambiguous we desire the encoder to predict variance in the latent space.
However given this ambiguity in latent space the reconstruction loss is minimized when the decoder averages over plausible shapes. 
Replacing these latent box features during training removes some ambiguity so that the reconstruction loss is minimized when the decoder produces a specific object without as much blur.


\textbf{Normalizing flow:}
The normalizing flow transforms the box features into the distribution $\normal(0, 1)$ of the VAE prior, providing two important properties.
First, this maps the arbitrary range of the box features into the correct range for sampling from the VAE without requiring a distance function in latent box space.
Second, because the normalizing flow tends to be locally smooth, uncertainty in latent-box space corresponds to rotation, translation, and resizing uncertainty of the bounding box, allowing the VAE prior to model the variance of our dataset.



\textbf{Computing the Plausible Set:} 
$ICP$ finds local, not global, minima and typically $ICP$ is run many times with different initializations.
Our dataset $\dataset_{TEST}$ contains many copies of each object at different rotations, and these copies serve the function of different initializations.
However, there are some limitations of our plausible set computation.
Our algorithm to compute $\plausibles$ has quadratic complexity which
limits the size of the test dataset.
In addition, our observation model explicitly ignores small depth errors without considering correlation of errors between pixels, yet small but correlated depth differences could be used to identify larger shapes.
Similarly, we explicitly discard depth values on the borders of shapes as independently these pixels tend to be noisy, yet again correlated depth values may provide useful information that is observable even with the independent noise.
Our network $\shapecompletion$ may use such features, but $\plausibles$ will not, thus our evaluation may be overly harsh on our network, penalizing it for not generating shapes in $\plausibles$ even when they are not plausible.

%% file: sections/conclusion.tex

\section{Conclusion}

In this work, we proposed $\ourmethod$, a method for generating diverse yet plausible 3D completions of a 2.5D depth image.
A normalizing flow transforms the side information of the true shape bounding box into a feature space, which is used during training to encourage an encoder to generate diverse latent space samples, and to aid the decoder in producing plausible samples.
To evaluate this method on a specific dataset we proposed a shape matching method to generate a set of plausible completions, as well as metrics for plausible diversity.
In experiment $\ourmethod$ generated diverse samples and outperformed existing approaches for depth images with ambiguous reconstructions. 

%% file: appendix/ExtendedExperiments.tex

\section{Further Experiment details}

\subsection{Generalizing of shape}

\begin{figure}[ht]
    \centering
    \includegraphics[width=0.99\textwidth]{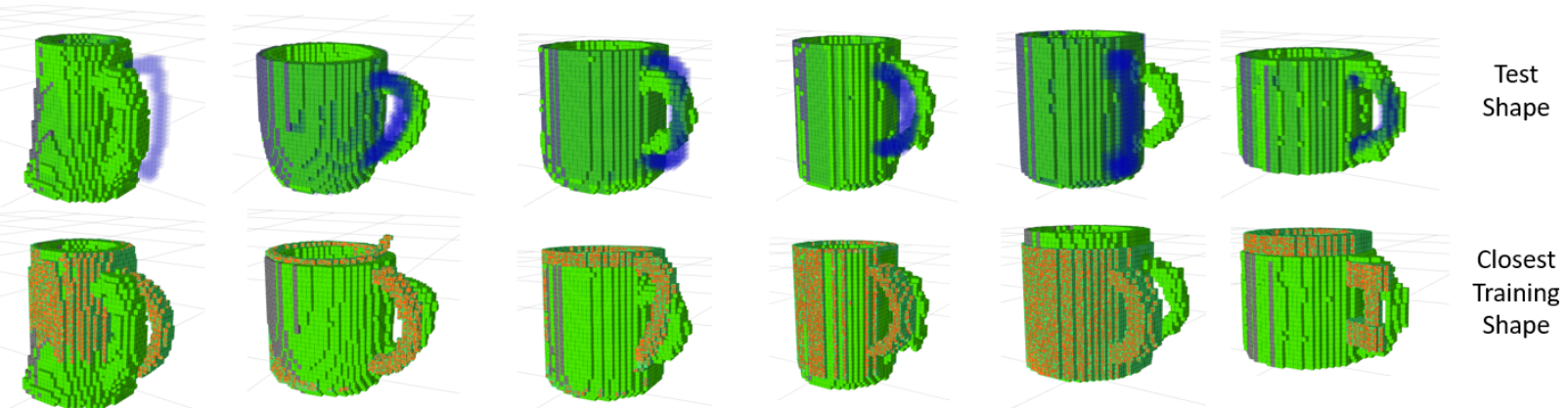}
    \caption{Similarity to the Training shapes. Completions (Green) from a depth view (grey) are shown in both the top and bottom row. 
    In all cases the handle is occluded from the depth view.
    The top row additionally shows ground truth (transparent blue), while the bottom row additionally shows the training shape (orange) that is closest (chamfer distance) to the ground truth test shape. The completed handle is sometimes closer to the training shape (e.g. left-most), sometimes closer to the test shape (e.g. right-most) and sometimes different from both.
    }
    \label{fig:train_similarity}
\end{figure}
In addition to our main contribution, we ask if these shape completion networks are ``completing the shape" or ``looking up the closest object from the training set". 
To evaluate this we examine the quality of the completions of the test shape as compared to the \textit{training} shapes.
For each test shape from the Shapenet Mugs dataset we compute the closest (chamfer distance) training shape.
We then sample 10 completions from PSSNet using the 2.5D view of the test shape and compute chamfer distance to both the closest-training and test shapes. 
Over all 26640 samples, the average chamfer distance to the test and closest-train shapes are 2.4mm and 3.8mm respectively. 
We find that in 2599 (approx 10\%) of samples the completion is closer to the training shape.
Numerically this indicates PSSNet (and presumably other shape completion networks) are more than searching for the nearest shape.

Qualitatively we notice features, such as the mug handle, sometimes visually appear closer to the closest-training shape. 
We visualize selected instances in Figure \ref{fig:train_similarity}.
We note that visually these completions represent the diversity we desire, where the completion of an occluded handle can vary.



\subsection{Physical Robot Details} \label{sec:physical}
\begin{figure}
    \centering
    \includegraphics[width=0.99\linewidth]{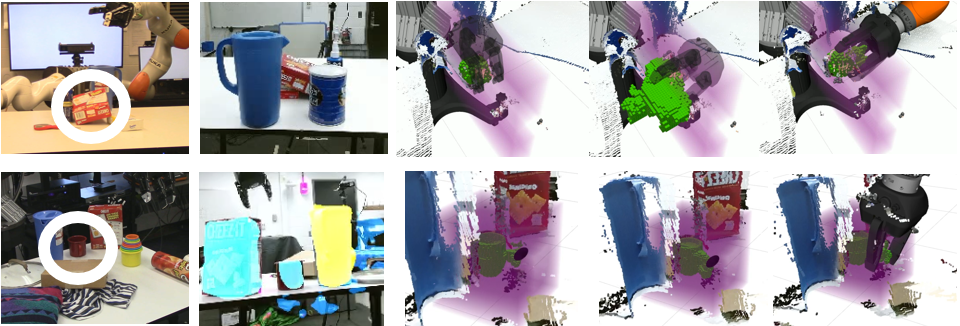}
    \caption{Robot scenarios for grasping Cheez-it box (top) and the mug (bottom).
    From left to right: The scene, the robot's view of the scene, 2 sampled completions using PSSNet, and a grasp attempt.}
    \label{fig:robot_appendix}
\end{figure}

We constructed two grasping scenarios on a physical robot, shown in \figref{fig:robot_appendix} and the accompanying video.
The points from a Kinect were filtered using an image segmenting network to construct known occupied and known free voxelgrids for the target object.
20 completions were sampled from which grasp poses were calculated, and then a grasp was attempted.
Our grasping strategies described below are simple but still serve to demonstrate the value of a diverse belief over shapes under ambiguity.

In the mug scenario, kinematics limits and clutter forced the robot to grasp the mug from the occluded handle on the far side from the robot.
For each completion a grasp was chosen with a handcoded orientation and grasp point as the furthest back possible grasp to avoid collision with other clutter.
The grasp attempted was the average of all valid grasp points with gripper width wide enough to capture all grasp points.
VAE-GAN sampled completions that did not have visible handles, resulting in most grasp poses in collision with other clutter.
Occasionally stray voxels appeared in VAE-GAN completions that generated valid grasp poses, but when attempted these grasps were not successful.
Using $\ourmethod$ sampled completions with handles generated valid grasps, which when executed resulted in successful grasps of the true mug.

In the Cheez-it scenario, clutter occluded all but a narrow slit from which only a small portion of the box was visible.
Potential grasps were sampled from both a top and side orientation with grasp point at the centroid of the completed object.
Completions from VAE-GAN were consistent, but the completed box was too shallow such that it appeared a top grasp would always be successful.
These attempted top grasps were unsuccessful because the gripper collided with the larger-than-expected box.
$\ourmethod$ again showed diversity with some completions thin and narrow and some as deep as the true box, so that it was unclear if a top grasp would be successful and thus the robot attempted and succeeded at side grasps.
